%% file: hcnn_arxiv.tex
\documentclass{article}

\newif\ifarxiv
\arxivtrue

\input{include/hcnn/packages.tex}
\usepackage[font={small}]{caption}
\usepackage{arydshln}

\input{include/hcnn/math_commands.tex}

\input{include/hcnn/title.tex}
\input{include/hcnn/authors.tex}

\iclrfinalcopy
\begin{document}

\maketitle

\begin{abstract}
\input{include/hcnn/sec0_abstract.tex}
\end{abstract}

\input{include/hcnn/sec1_intro.tex}
\input{include/hcnn/sec2_related.tex}
\input{include/hcnn/sec3_method.tex}
\input{include/hcnn/sec4_experiments.tex}
\input{include/hcnn/sec5_conclusion.tex}

\bibliography{hcnn_arxiv}
\bibliographystyle{iclr2019_conference}

\input{include/hcnn/appendix.tex}

\end{document}

%% file: include/hcnn/packages.tex
\usepackage{iclr2019_conference,times}
\definecolor{darkblue}{rgb}{0.0,0.0,0.55}
\usepackage[
   pagebackref=false,
   breaklinks=true,
   letterpaper=true,
   colorlinks,
   citecolor=darkblue,
   bookmarks=false]{hyperref}
\usepackage{graphicx}
\usepackage{enumitem}
\usepackage{color}
\usepackage{bbm}
\usepackage{mathtools}
\usepackage{subcaption}
\usepackage{url}
\usepackage{multirow}

%% file: include/hcnn/math_commands.tex
\usepackage{amsmath,amsfonts,bm}

\def\eqref#1{equation~\ref{#1}}

\def\1{\bm{1}}

\def\mA{{\bm{A}}}

\def\mI{{\bm{I}}}

\DeclareMathAlphabet{\mathsfit}{\encodingdefault}{\sfdefault}{m}{sl}
\SetMathAlphabet{\mathsfit}{bold}{\encodingdefault}{\sfdefault}{bx}{n}
\newcommand{\tens}[1]{\bm{\mathsfit{#1}}}

\def\tT{{\tens{T}}}
\def\tU{{\tens{U}}}

\def\tW{{\tens{W}}}
\def\tX{{\tens{X}}}



%% file: include/hcnn/title.tex
\title{Learning Implicitly Recurrent CNNs Through Parameter Sharing}

%% file: include/hcnn/authors.tex
\author{%
Pedro Savarese\\%
TTI-Chicago\\%
\texttt{savarese@ttic.edu}\\%
\And%
Michael Maire\\%
University of Chicago\\%
\texttt{mmaire@uchicago.edu}%
}

%% file: include/hcnn/sec0_abstract.tex
We introduce a parameter sharing scheme, in which different layers of a
convolutional neural network (CNN) are defined by a learned linear combination
of parameter tensors from a global bank of templates.  Restricting the number
of templates yields a flexible hybridization of traditional CNNs and recurrent
networks.  Compared to traditional CNNs, we demonstrate substantial parameter
savings on standard image classification tasks, while maintaining accuracy.

\vspace{0.5em}
Our simple parameter sharing scheme, though defined via soft weights, in
practice often yields trained networks with near strict recurrent structure;
with negligible side effects, they convert into networks with actual loops.
Training these networks thus implicitly involves discovery of suitable
recurrent architectures.  Though considering only the design aspect of
recurrent links, our trained networks achieve accuracy competitive with those
built using state-of-the-art neural architecture search (NAS) procedures.

\vspace{0.5em}
Our hybridization of recurrent and convolutional networks may also represent
a beneficial architectural bias.  Specifically, on synthetic tasks which are
algorithmic in nature, our hybrid networks both train faster and extrapolate
better to test examples outside the span of the training set.

%% file: include/hcnn/sec1_intro.tex
\section{Introduction}
\label{sec:intro}

The architectural details of convolutional neural networks (CNNs) have
undergone rapid exploration and improvement via both human hand-design~\citep{
vgg,
googlenet,
resnet1,
densenet,
sparsenet}
and automated search methods~\citep{
nas,
nas_progressive}.
Yet, this vast array of work limits itself to a circuit-like view of neural
networks.  Here, a CNN is regarded as a fixed-depth feed-forward circuit, with
a distinct parameter governing each internal connection.  These circuits are
often trained to perform tasks which, in a prior era, might have been (less
accurately) accomplished by running a traditional computer program coded by
humans.  Programs, and even traditional hardware circuits, have a more reusable
internal structure, including subroutines or modules, loops, and associated
control flow mechanisms.

We bring one aspect of such modularity into CNNs, by making it possible to
learn a set of parameters that is reused across multiple layers at different
depths.  As the pattern of reuse is itself learned, our scheme effectively
permits learning the length (iteration count) and content of multiple loops
defining the resulting CNN.  We view this approach as a first step towards
learning neural networks with internal organization reminiscent of computer
programs.  Though we focus solely on loop-like structures, leaving subroutines
and dynamic control flow to future work, this simple change suffices to yield
substantial quantitative and qualitative benefits over the standard baseline
CNN models.

While recurrent neural networks (RNNs) possess a loop-like structure by
definition, their loop structure is fixed a priori, rather than learned as part
of training.  This can actually be a disadvantage in the event that the
length of the loop is mismatched to the target task.  Our parameter sharing
scheme for CNNs permits a mix of loops and feed-forward layers to emerge.
For example, trained with our scheme, a 50-layer CNN might learn a 2-layer
loop that executes 5 times between layers 10 and 20, a 3-layer loop that runs
4 times from layers 30 to 42, while leaving the remaining layers to assume
independent parameter sets.  Our approach generalizes both CNNs and RNNs,
creating a hybrid.

\input{include/hcnn/fig_sharing_scheme.tex}

Figure~\ref{fig:sharing_scheme} diagrams the parameter sharing scheme
facilitating this hybridization.  Inspired by dictionary learning, different
network layers share, via weighted combination, global parameter templates.
This re-parameterization is fully differentiable, allowing learning of sharing
weights and template parameters.  Section~\ref{sec:method} elaborates, and also
introduces tools for analyzing learned loop structures.

Section~\ref{sec:experiments} demonstrates advantages of our hybrid CNNs across
multiple experimental settings.  Taking a modern CNN design as a baseline, and
re-parameterizing it according to our scheme improves:

\begin{itemize}
   \item{
      \textbf{Parameter efficiency.}
         Here, we experiment with the standard task of image classification
         using modern residual networks~\citep{resnet1,wide}.  This task is
         a good proxy for general usefulness in computer vision, as
         high-performance classification architectures often serve as a
         backbone for many other vision tasks, such as semantic
         segmentation~\citep{DeepLab,PSPNet}.

         Our parameter sharing scheme drastically reduces the number of
         unique parameters required to achieve a given accuracy on
         CIFAR~\citep{cifar} or ImageNet~\citep{imagenet} classification tasks.
         Re-parameterizing a standard residual network with our scheme cuts
         parameters, without triggering any drop in accuracy.  This suggests
         that standard CNNs may be overparameterized in part because, by design
         (and unlike RNNs), they lack capacity to learn reusable internal
         operations.
   }
   \item{
      \textbf{Extrapolation and generalization.}
         Here, we explore whether our hybrid networks expand the class of tasks
         that one can expect to train neural networks to accomplish.  This
         line of inquiry, focusing on synthetic tasks, shares motivations with
         work on Neural Turing Machines~\citep{ntm}.  Specifically, we would
         like neural networks to be capable of learning to perform tasks for
         which there are concise traditional solution algorithms.  \cite{ntm}
         uses sorting as an example task.  As we examine an extension of CNNs,
         our tasks take the form of queries about planar graphs encoded as
         image input.

         On these tasks, we observe improvements to both generalization ability
         and learning speed for our hybrid CNNs, in comparison to standard CNNs
         or RNNs.  Our parameter sharing scheme, by virtue of providing an
         architectural bias towards networks with loops, appears to assist in
         learning to emulate traditional algorithms.
   }
\end{itemize}

An additional side effect, seen in practice in many of our experiments, is that
two different learned layers often snap to the same parameter values.  That is,
layers $i$ and $j$, learn coefficient vectors $\boldsymbol{\alpha}^{(i)}$ and
$\boldsymbol{ \alpha}^{(j)}$ (see Figure~\ref{fig:sharing_scheme}) that
converge to be the same (up to scaling).  This is a form of architecture
discovery, as it permits representation of the CNN as a loopy wiring diagram
between repeated layers.  Section~\ref{sec-imp_rec} presents example results.
We also draw comparisons to existing neural architecture search (NAS)
techniques.  By simply learning recurrent structure as byproduct of training
with standard stochastic gradient descent, we achieve accuracy competitive
with current NAS procedures.

Before delving into the details of our method, Section~\ref{sec:related}
provides additional context in terms of prior work on recurrent models,
parameter reduction techniques, and program emulation.
Sections~\ref{sec:method} and~\ref{sec:experiments} describe our hybrid
shared-parameter CNN, experimental setup, and results.
Section~\ref{sec:conclusion} concludes with commentary on our results and
possible future research pathways.\footnote{Our code is available at
\url{https://github.com/lolemacs/soft-sharing}}

%% file: include/hcnn/fig_sharing_scheme.tex
\begin{figure}
   \begin{center}
      \includegraphics[width=0.9\textwidth]{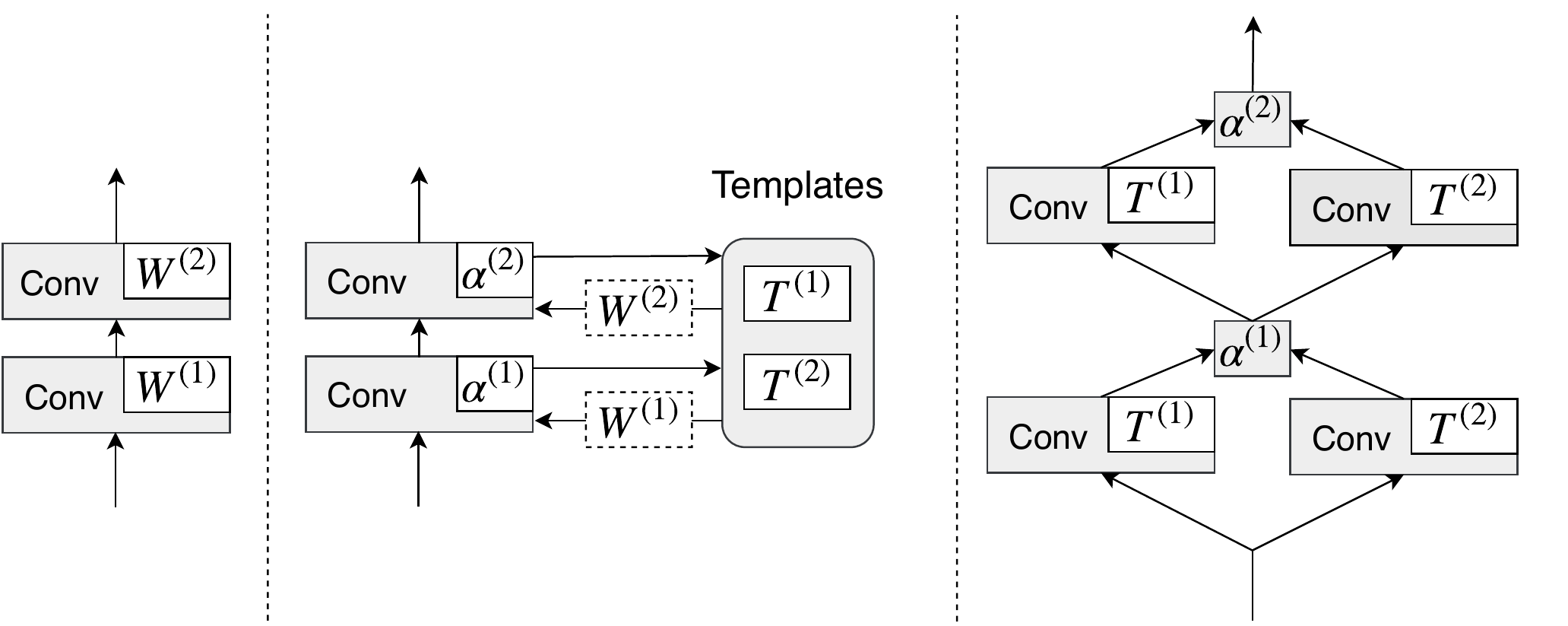}
   \end{center}
   \caption{
      Parameter sharing scheme.
      \emph{\textbf{Left:}}
         A CNN (possibly a variant such as a residual network), with each
         convolutional layer $i$ containing an individual parameter set
         $\tW^{(i)}$.
      \emph{\textbf{Middle:}}
         Parameter sharing among layers, where parameter templates
         $\tT^{(1)}, \tT^{(2)}$ are shared among each layer $i$, which now only
         contains a $2$-dimensional parameter $\boldsymbol{\alpha}^{(i)}$.
         Weights $\tW^{(i)}$ (no longer parameters, illustrated with dotted
         boxes) used by layer $i$ are generated from
         $\boldsymbol{\alpha}^{(i)}$ and templates $\tT^{(1)}, \tT^{(2)}$.
      \emph{\textbf{Right:}}
         If weights $\tW^{(i)}$ are outputs of a linear function (as in our
         method), learning parameter templates can be viewed as learning layer
         templates, offering a new (although equivalent) perspective for the
         middle diagram.  Non-linearities are omitted for simplicity.
   }
   \label{fig:sharing_scheme}
\end{figure}

%% file: include/hcnn/sec2_related.tex
\section{Related Work}
\label{sec:related}

Recurrent variants of CNNs are used extensively for visual tasks.  Recently,
\cite{feedback} propose utilizing a convolutional LSTM~\citep{convLSTM} as a
generic feedback architecture.  RNN and CNN combinations have been used for
scene labeling~\citep{pinheiro}, image captioning with
attention~\citep{showattendtell}, and understanding video~\citep{ltrcnn},
among others.  These works combine CNNs and RNNs at a coarse scale, and in a
fixed hand-crafted manner.  In contrast, we learn the recurrence structure
itself, blending it into the inner workings of a CNN.

Analysis of residual networks~\citep{resnet1} reveals possible connections to
recurrent networks stemming from their design~\citep{resnet_cortex}.
\cite{unrolled} provide evidence that residual networks learn to iteratively
refine feature representations, making an analogy between a very deep residual
network and an unrolled loop.  \cite{resnet_iter} further explore this
connection, and experiment with training residual networks in which some
layers are forced to share identical parameters.  This hard parameter sharing
scheme again builds a predetermined recurrence structure into the network.  It
yields successfully trained networks, but does not exhibit the type of
performance gains that Section~\ref{sec:experiments} demonstrates for our
soft parameter sharing scheme.

Closely related to our approach is the idea of
hypernetworks~\citep{hypernetworks}, in which one part of a neural network is
parameterized by another neural network.  Our shared template-based
re-parameterization could be viewed as one simple choice of hypernetwork
implementation.  Perhaps surprisingly, this class of ideas has not been well
explored for the purpose of reducing the size of neural networks.  Rather,
prior work has achieved parameter reduction through explicit representation
bottlenecks~\citep{squeezenet}, sparsifying connection structure
\citep{deepexpander,condensenet,sparsenet}, and pruning trained networks
\citep{han2015deep_compress}.

Orthogonal to the question of efficiency, there is substantial interest in
extending neural networks to tackle new kinds of tasks, including emulation
of computer programs.  Some approach this problem using additional supervision
in the form of execution traces~\citep{npi,cai2017}, while other focus on
development of network architectures that can learn from input-output pairs
alone~\citep{ntm,dnc,zaremba2016,neuralalu}.  Our experiments on synthetic
tasks fall into the latter camp.  At the level of architectural strategy,
\cite{neuralalu} benefit from changing the form of activation function to bias
the network towards correctly extrapolating common mathematical formulae.  We
build in a different implicit bias, towards learning iterative procedures
within a CNN, and obtain a boost on correctly emulating programs.

%% file: include/hcnn/sec3_method.tex
\section{Soft Parameter Sharing}
\label{sec:method}

In convolutional neural networks (CNNs) and variants such as residual CNNs
(ResNets)~\citep{resnet1} and DenseNets~\citep{densenet}, each convolutional
layer $i$ contains a set of parameters $\tW^{(i)}$, with no explicit relation
between parameter sets of different layers.  Conversely, a strict structure is
imposed to layers of recurrent neural networks (RNNs), where, in standard
models~\citep{lstm}, a single parameter set $\tW$ is shared among all time
steps.  This leads to a program-like computational flow, where RNNs can be seen
as loops with fixed length and content.  While some RNN variants~\citep{
stackedrnn,clockwork,densenetclique} are less strict on the length or content
of loops, these are still typically fixed beforehand.

As an alternative to learning hard parameter sharing schemes -- which
correspond to the strict structure present in RNNs -- our method consists of
learning soft sharing schemes through a relaxation of this structure.  We
accomplish this by expressing each layer's parameters $\tW^{(i)}$ as a linear
combination of parameter templates $\tT^{(1)}, \dots, \tT^{(k)}$, each with
the same dimensionality as $\tW^{(i)}$:
\begin{equation}
   \tW^{(i)} \coloneqq \sum_{j=1}^k \alpha^{(i)}_j \tT^{(j)}
   \label{sharing1}
\end{equation}
where $k$ is the number of parameter templates (chosen freely as a
hyperparameter) and $\boldsymbol{\alpha}^{(i)}$, a $k$-dimensional vector, is
the coefficients of layer $i$.  Figure~\ref{fig:sharing_scheme} (left and
middle) illustrates the difference between networks trained with and without
our method.  This relaxation allows for coefficients and parameter templates to
be (jointly) optimized with gradient-based methods, yielding negligible extra
computational cost, with a single constraint that only layers with same
parameter sizes can share templates.  Note that constraining coefficients
$\boldsymbol{\alpha}^{(i)}$ to be one-hot vectors leads to hard sharing
schemes, at the cost of non-differentiability.

Having $k$ as a free parameter decouples the number of parameters in network
from its depth.  Typically, $L$ convolutional layers with constant channel and
kernel sizes $C, K$ have $O(L C^2 K^2)$ total parameters.  Our soft sharing
scheme changes the total number of parameters to
$O(kL + kC^2 K^2) = O(kC^2 K^2)$.  Sections~\ref{sec-cifar}
and~\ref{sec-imagenet} show that we can decrease the parameter count of
standard models without significantly impacting accuracy, or simply attain
higher accuracy with $k=L$.

In the next two subsections, we discuss two consequences of the linearity of
Equation~(\ref{sharing1}).  First, it enables alternative interpretations of
our method.  Second, and a major advantage, as is the case in many linear
relaxations of integer problems, we are able to extract hard sharing schemes
in practice, and consequently detect implicit self-loops in a CNN trained with
our method.

\input{include/hcnn/fig_implicit_recurrence.tex}

\subsection{Interpretation}

For layers $i$ that are linear in $\tW^{(i)}$ (\emph{e.g.}~matrix
multiplication, convolution), we can view our method as learning template
layers which are shared among a network.  More specifically, for a
convolutional layer $\tU^{(i)}(\tX) = \tW^{(i)} * \tX$, and considering
Equation~(\ref{sharing1}):
\begin{equation}
\begin{split}
   \tU^{(i)}(\tX) &=
      \tW^{(i)} * \tX = \sum_{j=1}^k \alpha^{(i)}_j \tT^{(j)} * \tX
\end{split}
\label{templatelayer}
\end{equation}
where $\tT^{(j)} * \tX$, the result of a convolution with filter sets
$\tT^{(j)}$, can be seen as the output of a template layer with individual
parameters $\tT^{(j)}$.  Such layers can be seen as global feature extractors,
and coefficients $\boldsymbol{\alpha}^{(i)}$ determine which features are
relevant for the $i$'th computation of a network.  This is illustrated in
Figure~\ref{fig:sharing_scheme} (right diagram).

This view gives a clear connection between coefficients $\boldsymbol{\alpha}$
and the network's structure.
Having
$\boldsymbol{\alpha}^{(i)} = \boldsymbol{\alpha}^{(i+2)}$
yields
$\tW^{(i)} =
   \sum_{j=1}^k \alpha^{(i)}_j \tT^{(j)} =
   \sum_{j=1}^k \alpha^{(i+2)}_j \tT^{(j)} =
   \tW^{(i+2)}$,
and hence layers $i$ and $i+2$ are functionally equivalent.  Such a network
can be folded to generate an equivalent model with two layers and a self-loop,
an explicitly recurrent network.  While this is also possible for networks
without parameter sharing, a learned alignment of $C^2 K^2$ parameters is
required (unlikely in practice), instead of aligning only $k \leq L$
parameters.

\subsection{Implicit Recurrences}

To identify which layers in a network perform approximately the same operation,
we can simply check whether their coefficients are similar.  We can condense
this information for all pairs of layers $i,j$ in a similarity matrix $S$,
where $S_{i,j} = s(\boldsymbol{\alpha}^{(i)}, \boldsymbol{\alpha}^{(j)})$ for
some similarity measure $s$.

For networks with normalization layers, the network's output is invariant to
weight rescaling.  In this setting, a natural measure is
$s(\boldsymbol{\alpha}^{(i)}, \boldsymbol{\alpha}^{(j)}) =
   \frac{| \langle \boldsymbol{\alpha}^{(i)}, \boldsymbol{\alpha}^{(j)} \rangle |}
      {\lVert \boldsymbol{\alpha}^{(i)} \rVert \lVert \boldsymbol{\alpha}^{(j)} \rVert}$
(absolute value of cosine similarity), since it possess this same
property.\footnote{We take the absolute value for simplicity: while negating a
layer's weights can indeed impact the network's output, this is circumvented by
adding a $-1$ multiplier to, for example, the input of layer $i$ in case
$\langle \boldsymbol{\alpha}^{(i)}, \boldsymbol{\alpha}^{(j)} \rangle$
is negative, along with
$\boldsymbol{\alpha}^{(i)} \gets - \boldsymbol{\alpha}^{(i)}$.}
We call $S$ the layer similarity matrix (LSM).
Figure~\ref{fig:implicit_recurrence} illustrates and Section~\ref{sec-imp_rec}
shows experimentally how it can be used to extract recurrent loops from trained
CNNs.

While structure might emerge naturally, having a bias towards more structured
(recurrent) models might be desirable.  In this case, we can add a
\emph{recurrence regularizer} to the training objective, pushing parameters to
values which result in more structure.  For example, we can add the negative
of sum of elements of the LSM:
   $\mathcal L_{R} = \mathcal L - \lambda_R \sum_{i,j} S_{i,j}$,
where $\mathcal L$ is the original objective.  The larger $\lambda_R$ is, the
closer the elements of $S$ will be to $1$.  At an extreme case, this
regularizer will push all elements in $S$ to $1$, resulting in a network with
a single layer and a self-loop.

%% file: include/hcnn/fig_implicit_recurrence.tex
\begin{figure}
   \centering
   \includegraphics[trim={0 1.1cm 0 0},clip,width=0.9\textwidth]{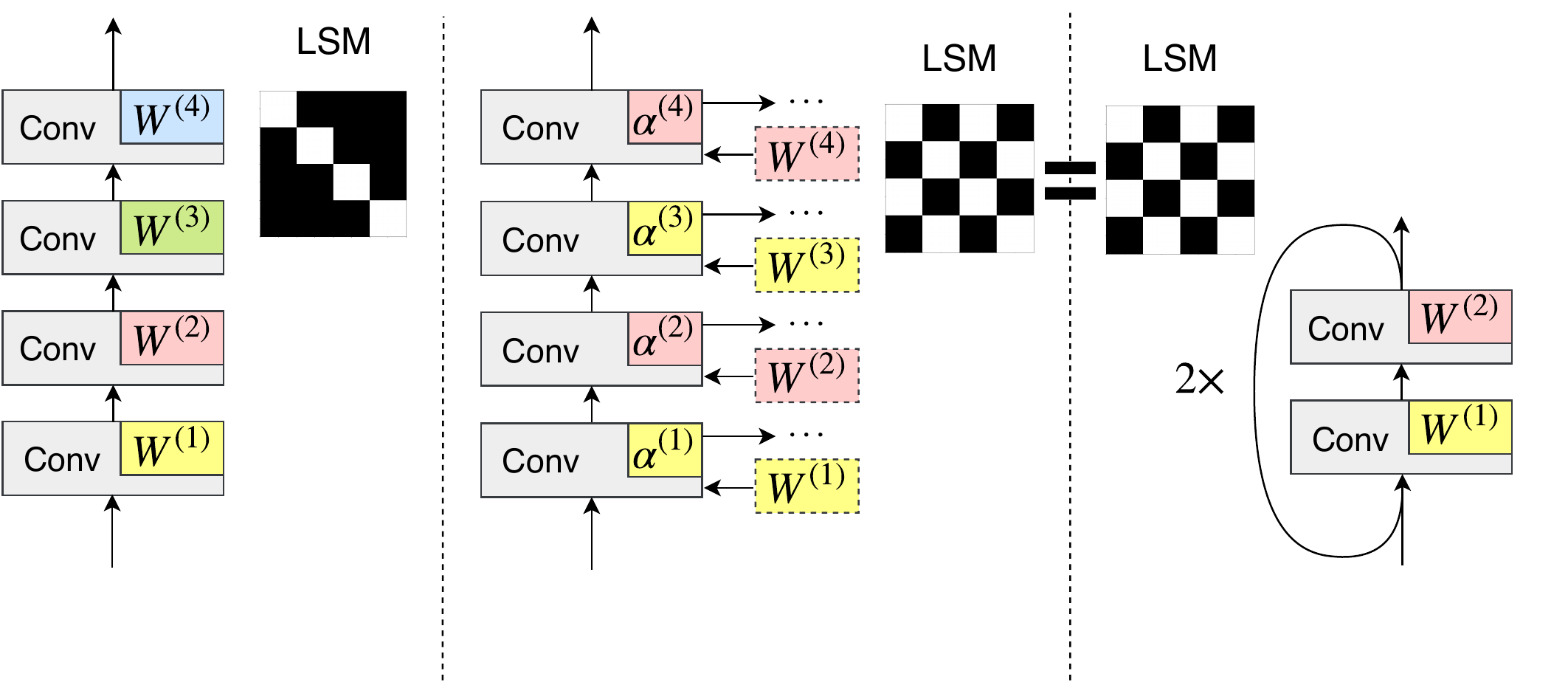}
   \caption{
      Connection between the LSM matrix $S$
      \big(
         where $S_{i,j} =
            \frac{| \langle \boldsymbol{\alpha}^{(i)}, \boldsymbol{\alpha}^{(j)} \rangle |}
                 {\lVert \boldsymbol{\alpha}^{(i)} \rVert \lVert \boldsymbol{\alpha}^{(j)} \rVert}$
      \big)
      and the structure of the network.  White and black entries correspond to
      maximum and minimum similarities ($S_{i,j} = 1$ and $S_{i,j} = 0$,
      respectively).
      \emph{\textbf{Left:}}
         Empirically, CNNs present no similarity between parameters of
         different layers.
      \emph{\textbf{Middle:}}
         Trained with our method, the layer similarity matrix (LSM) captures
         similarities between different layers, including pairs with close to
         maximum similarity.  Such pairs (depicted by same-colored coefficients
         and weights, and by white entries in the LSM) perform similar
         operations on their inputs.
      \emph{\textbf{Right:}}
         We can tie together parameters of similar layers, creating a hard
         parameter sharing scheme.  The network can then be folded, creating
         self-loops and revealing an explicit recurrent computation structure.
      }
   \label{fig:implicit_recurrence}
\end{figure}

%% file: include/hcnn/sec4_experiments.tex
\section{Experiments}
\label{sec:experiments}

We begin by training variants of standard models with soft parameter sharing,
observing that it can offer parameter savings with little impact on
performance, or increase performance at the same parameter count.
Section~\ref{sec-imp_rec} demonstrates conversion of a trained model into
explicitly recurrent form.  We then examine synthetic tasks
(Section~\ref{sec-rec_task}), where parameter sharing improves generalization.
Appendix~\ref{appendix2} contains details on the initialization for the
coefficients $\boldsymbol{\alpha}$.

\subsection{Classification on CIFAR}
\label{sec-cifar}

\input{include/hcnn/tab_cifar.tex}

\input{include/hcnn/fig_cifar.tex}

The CIFAR-10 and CIFAR-100 datasets~\citep{cifar} are composed of $60,000$
colored $32 \times 32$ images, labeled among $10$ and $100$ classes
respectively, and split into $50,000$ and $10,000$ examples for training and
testing.  We pre-process the training set with channel-wise normalization, and
use horizontal flips and random crops for data augmentation,
following~\cite{resnet1}.

Using Wide ResNets (WRN)~\citep{wide} as a base model, we train networks with
the proposed soft parameter sharing method.  Since convolution layers have
different number of channels and kernel sizes throughout the network, we
create 3 layer groups and only share templates among layers in the same group.
More specifically, WRNs for CIFAR consist of 3 stages whose inputs and outputs
mostly have a constant number of channels ($C$, $2C$ and $4C$, for some $C$).
Each stage contains $\frac{L-4}{3}$ layers for a network with depth $L$, hence
we group layers in the same stage together, except for the first two, a
residual block whose input has a different number of channels.

Thus, all layers except for the first 2 in each stage perform parameter
sharing (illustrated in left diagram of Figure~\ref{fig:swrn_folding}).
Having $k$ templates per group means that $\frac{L-4}{3} - 2$ convolution
layers share $k$ parameter templates.  We denote by SWRN-$L$-$w$-$k$ a WRN
with $L$ layers, widen factor $w$ and $k$ parameter templates per group
(trained with our method).  Setting $k = \frac{L-4}{3} - 2$ means we have one
parameter template per layer, and hence no parameter reduction.  We denote
SWRN-$L$-$w$ (thus omitting $k$) as a model in this setting.

Following~\cite{wide}, we train each model for $200$ epochs with SGD and
Nesterov momentum of $0.9$ and a batch size of $128$.  The learning rate is
initially set to $0.1$ and decays by a factor of $5$ at epochs 60, 120 and 160.
We also apply weight decay of $5\times 10^{-4}$ on all parameters except for
the coefficients $\boldsymbol{\alpha}$.

Tables~\ref{c10} and~\ref{c10_2} present results.  Networks trained with our
method yield superior performance in the setting with no parameter reduction:
SWRN 28-10 presents $6.5\%$ and $2.5\%$ lower relative test errors on C-10 and
C-100, compared to the base WRN 28-10 model.  With fewer templates than layers,
SWRN 28-10-1 (all 6 layers of each group perform the same operation), performs
virtually the same as the base WRN 28-10 network, while having $\frac{1}{3}$
of its parameters.  On CIFAR-10, parameter reduction ($k=2$) is beneficial to
test performance: the best performance is achieved by SWRN 28-18-2
({3.43\% test error}), outperforming the ResNeXt-29 16x64 model~\citep{
resnext}, while having fewer parameters (55M against 68M) and no bottleneck
layers.

Figure~\ref{fig:cifar} shows that our parameter sharing scheme uniformly
improves accuracy-parameter efficiency; compare the WRN model family
(solid red) to our SWRN models (dotted red).

Table~\ref{c10_nas} presents a comparison between our method and neural
architecture search (NAS) techniques \citep{nas,snas,darts,enas,amoeba} on
CIFAR-10 -- results differ from Table~\ref{c10_2} solely due to
cutout~\citep{cutout}, which is commonly used in NAS literature; NAS results
are quoted from their respective papers.  Our method outperforms architectures
discovered by recent NAS algorithms, such as DARTS~\citep{darts},
SNAS~\citep{snas} and ENAS~\citep{enas}, while having similarly low training
cost.  We achieve $2.69\%$ test error after training less than $10$ hours on a
single NVIDIA GTX 1080 Ti.  This accuracy is only bested by NAS techniques
which are several orders of magnitude more expensive to train.  Being
based on Wide ResNets, our models do, admittedly, have more parameters.

Comparison to recent NAS algorithms, such as DARTS and SNAS, is particularly
interesting as our method, though motivated differently, bears some notable
similarities.  Specifically, all three methods are gradient-based and use an
extra set of parameters (architecture parameters in DARTS and SNAS) to perform
some kind of soft selection (over operations/paths in DARTS/SNAS; over
templates in our method).  As Section \ref{sec-imp_rec} will show, our learned
template coefficients $\boldsymbol{\alpha}$ can often be used to transform our
networks into an explicitly recurrent form - a discovered CNN-RNN hybrid.

To the extent that our method can be interpreted as a form of architecture
search, it might be complementary to standard NAS methods.  While NAS methods
typically search over operations (\emph{e.g.} activation functions;
$3\times3$ or $5\times5$ convolutions; non-separable, separable, or grouped
filters; dilation; pooling), our soft parameter sharing can be seen as a search
over recurrent patterns (which layer processes the output at each step).  These
seem like orthogonal aspects of neural architectures, both of which may be
worth examining in an expanded search space.  When using SGD to drive
architecture search, these aspects take on distinct forms at the implementation
level: soft parameter sharing across layers (our method) vs hard parameter
sharing across networks (recent NAS methods).

\input{include/hcnn/tab_imagenet_c10nas.tex}

\subsection{Classification on ImageNet} \label{sec-imagenet}

We use the ILSVRC 2012 dataset~\citep{imagenet} as a stronger test of our
method.  It is composed of $1.2$M training and $50,000$ validation images,
drawn from 1000 classes.  We follow~\cite{gross}, as in \cite{wide,densenet,
resnext}, and report Top-1 and Top-5 errors on the validation set using single
$224 \times 224$ crops.  For this experiment, we use WRN 50-2 as a base model,
and train it with soft sharing and no parameter reduction.  Having bottleneck
blocks, this model presents less uniform number of channels of layer inputs and
outputs.  To apply our method, we group convolutions in $12$ groups: for each
of the 4 stages in a WRN 50-2, we create 3 groups, one for each type of layer
in a bottleneck unit ($C \to B$, $B \to B$ and $B \to C$ channel mappings, for
bottleneck $B$).  Without any change in hyperparameters, the network trained
with our method outperforms the base model and also deeper models such as
DenseNets (though using more parameters), and performs close to ResNet-200, a
model with four times the number of layers and a similar parameter count.  See
Table~\ref{tab:imagenet}.

\subsection{Learning Implicit Recurrences} \label{sec-imp_rec}

Results on CIFAR suggest that training networks with few parameter templates
$k$ in our soft sharing scheme results in performance comparable to the base
models, which have significantly more parameters.  The lower $k$ is, the larger
we should expect the layer similarities to be: in the extreme case where $k=1$,
all layers in a sharing scheme have similarity $1$, and can be folded into a
single layer with a self-loop.

\input{include/hcnn/fig_swrn_folding.tex}

For the case $k > 1$, there is no trivial way to fold the network, as layer
similarities depend on the learned coefficients.  We can inspect the model's
layer similarity matrix (LSM) and see if it presents implicit recurrences: a
form of recurrence in the rows/columns of the LSM.  Surprisingly, we observe
that rich structures emerge naturally in networks trained with soft parameter
sharing, \emph{even without the recurrence regularizer}.
Figure~\ref{fig:swrn_folding} shows the per-stage LSM for CIFAR-trained
SWRN 28-10-4.  Here, the six layers of its stage-2 block can be folded into a
loop of two layers, leading to an error increase of only $0.02\%$.
Appendix~\ref{appendix} contains an additional example of network folding,
diversity of LSM patterns across different runs, and an epoch-wise evolution of
the LSM, showing that many patterns are observable after as few as 5 epochs of
training.

\subsection{Evaluation on Naturally Recurrent Tasks}
\label{sec-rec_task}

\input{include/hcnn/fig_shortest.tex}

While the propensity of our parameter sharing scheme to encourage learning of
recurrent networks is a useful parameter reduction tool, we would also like to
leverage it for qualitative advantages over standard CNNs.  On tasks for which
a natural recurrent algorithm exists, does training CNNs with soft parameter
sharing lead to better extrapolation?

To answer this, we set up a synthetic algorithmic task: computing shortest
paths.  Examples are $32 \times 32$ grids containing two query points and
randomly (with probability $0.1$) placed obstacles.  The objective is to
indicate which grid points belong to a shortest path between the query points.

We use curriculum learning for training, allowing us to observe how well each
model adapts to more difficult examples as training phases progress.  Moreover,
for this task curriculum learning causes faster learning and superior
performance for all trained models.

Training consists of 5 curriculum phases, each one containing 5000 examples.
The maximum allowed distance between the two query points increases at each
phase, thus increasing difficulty.  In the first phase, each query point is
within a $5 \times 5$ grid around the other query point, and the grid size
increases by $2$ on each side at each phase, yielding a final grid size of
$21 \times 21$ at phase 5.

We train a CNN, a CNN with soft parameter sharing and one template per layer
(SCNN), and an SCNN with recurrence regularizer $\lambda_R = 0.01$.  Each model
trains for 50 epochs per phase with Adam \citep{adam} and a fixed learning
rate of $0.01$.  As classes are heavily unbalanced and the balance itself
changes during phases, we compare $F_1$ scores instead of classification error.

Each model starts with a $1 \times 1$ convolution, mapping the 2 input channels
to 32 output channels.  Next, there are 20 channel-preserving $3 \times 3$
convolutions, followed by a final $1 \times 1$ convolution that maps 32
channels to 1.  Each of the 20 $3 \times 3$ convolutions is followed by batch
normalization~\citep{bn}, a ReLU non-linearity~\citep{relu}, and has a
1-skip connection.

Figure~\ref{fig:shortest} shows one example from our generated dataset and the
training curves for the 3 trained models: the SCNN not only outperforms the
CNN, but adapts better to harder examples at new curriculum phases.  The
SCNN is also advantaged over a more RNN-like model: with the recurrence
regularizer $\lambda_R = 0.01$, all entries in the LSM quickly converge $1$,
as in a RNN.  This leads to faster learning during the first phase, but
presents issues in adapting to difficulty changes in latter phases.

%% file: include/hcnn/tab_cifar.tex
\begin{table}[t]
\parbox{.49\linewidth}{
   \centering
   \caption{
      Test error ($\%$) on CIFAR-10 and CIFAR-100.
      SWRN 28-10, the result of training a WRN 28-10 with our method and one
      template per layer, significantly outperforms the base model, suggesting
      that our method aids optimization (both models have the same capacity).
      SWRN 28-10-1, with a single template per sharing group, performs close
      to WRN 28-10 while having significantly less parameters and capacity.
      * indicates models trained with dropout $p=0.3$ \citep{dropout}.
      Results are average of 5 runs.
   }
   \label{c10}
   \small
   \setlength{\tabcolsep}{8pt}
   \begin{tabular}{@{}l|c|c|c@{}}
   \textbf{CIFAR}   &  Params  & C-10+          & C-100+ \\ \hline
      WRN 28-10     &  36M     & 4.0            & 19.25 \\
      WRN 28-10*    &  36M     & 3.89           & 18.85 \\ \hline

      SWRN 28-10    &  36M     & \textbf{3.74}  & 18.78 \\
      SWRN 28-10*   &  36M     & 3.88           & \textbf{18.43}   \\
      SWRN 28-10-1  &  12M     & 4.01           & 19.73 \\
   \end{tabular}
}
\hfill
\parbox{.49\linewidth}{
   \centering
   \caption{
      Performance of wider SWRNs. Parameter reduction ($k=2$) leads to lower
      errors for CIFAR-10, with models being competitive against newer model
      families that have bottleneck layers, group convolutions, or many
      layers.  Best SWRN results are in bold, and best overall results are
      underlined.
   }
   \label{c10_2}
   \small
   \setlength{\tabcolsep}{8pt}
   \begin{tabular}{@{}l|c|c|c@{}}
   \textbf{CIFAR}       &  Params  & C-10+  & C-100+ \\ \hline
      ResNeXt-29 16x64  &  68M     & 3.58   & 17.31 \\
      DenseNet 100-24   &  27M     & 3.74   & 19.25 \\
      DenseNet 190-40   &  26M     & 3.46   & \underline{17.18} \\ \hline
      SWRN 28-10*       &  36M     & 3.88   & 18.43 \\
      SWRN 28-10-2*     &  17M     & 3.75   & 18.66 \\ \hdashline
      SWRN 28-14*       &  71M     & 3.67   & 18.25 \\
      SWRN 28-14-2*     &  33M     & 3.69   & 18.37 \\ \hdashline
      SWRN 28-18*       &  118M    & 3.48   & \textbf{17.43 } \\
      SWRN 28-18-2*     &  55M     & \underline{\textbf{3.43}}  & 17.75 \\
   \end{tabular}
}
\end{table}

%% file: include/hcnn/fig_cifar.tex
\begin{figure}[t]
   \centering
   \begin{minipage}[b]{0.49\linewidth}
      \centering
      \scriptsize{\textsf{CIFAR-10}}\\
      \includegraphics[trim={0 0 0 0.6cm},clip,width=\textwidth]{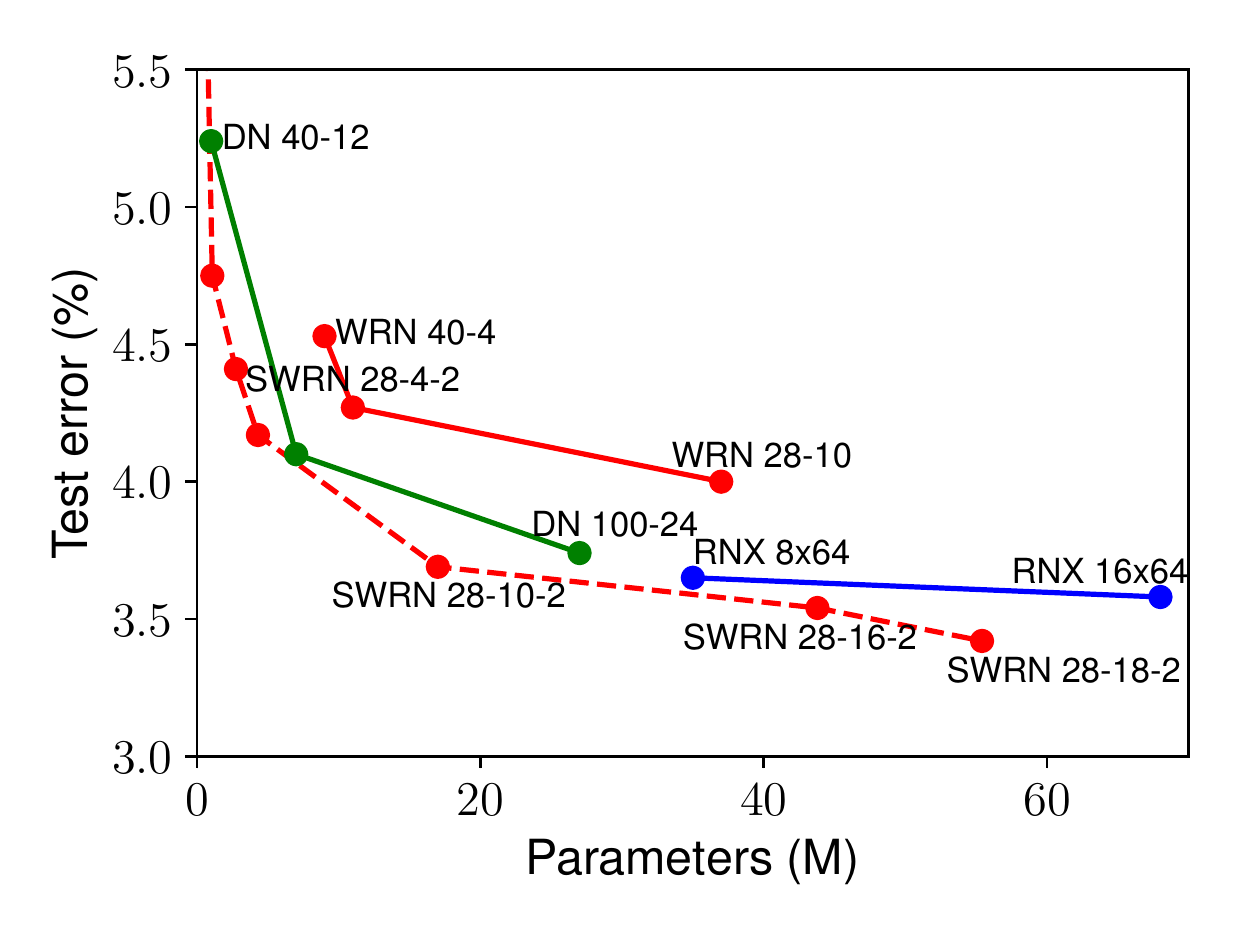}
   \end{minipage}
   \begin{minipage}[b]{0.49\linewidth}
      \centering
      \scriptsize{\textsf{CIFAR-100}}\\
      \includegraphics[trim={0 0 0 0.6cm},clip,width=\textwidth]{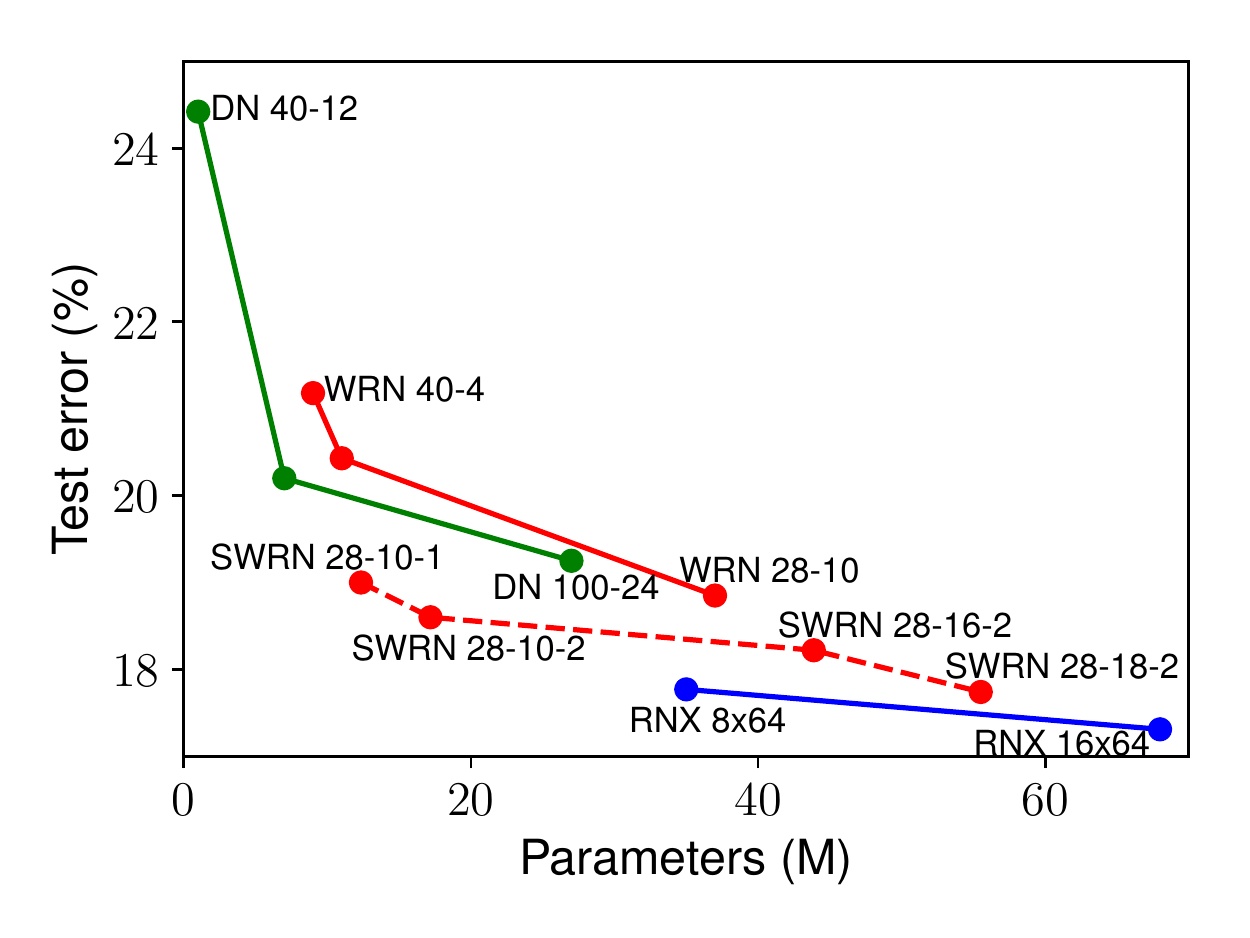}
   \end{minipage}
   \vspace{-7pt}
   \caption{
      Parameter efficiency for different models.  On both CIFAR-10 and
      CIFAR-100, SWRNs are significantly more efficient than WRNs.
      DN and RNX denotes DenseNet and ResNeXt, respectively, and are plotted
      for illustration: both models employ orthogonal efficiency techniques,
      such as bottleneck layers. Best viewed in color.
   }
   \label{fig:cifar}
\end{figure}

%% file: include/hcnn/tab_imagenet_c10nas.tex
\begin{table}
\parbox{.43\linewidth}{
   \centering
   \caption{
      \emph{(below)}
      ImageNet classification results: training WRN 50-2 with soft parameter
      sharing leads to better performance by itself, without any tuning on the
      number of templates $k$. Top-1 and Top-5 errors (\%) are computed using
      a single crop.
   }
   \label{tab:imagenet}
   \small
   \setlength{\tabcolsep}{8pt}
   \begin{tabular}{@{}l|c|c|c@{}}
   \textbf{ImageNet}  &  Params   & Top-1    & Top-5 \\ \hline
         WRN 50-2     &  69M      & 22.0     & 6.05 \\
         DenseNet-264 &  33M      & 22.15    & 6.12 \\
         ResNet-200   &  65M      & 21.66    & 5.79 \\
         SWRN 50-2    &  69M      & 21.74    & 5.95 \\
   \end{tabular}
   \vspace{0.06\linewidth}
   \caption{
      \emph{(right)}
      Test error ($\%$) on CIFAR-10 of SWRNs and models found via neural
      architecture search (NAS) (all trained with cutout). Networks trained
      with soft parameter sharing provide competitive performance against NAS
      methods while having low computational cost.
   }
   \label{c10_nas}
}
\hfill
\parbox{.54\linewidth}{
   \centering
   \small
   \vspace{-7pt}
   \setlength{\tabcolsep}{6pt}
   \begin{tabular}{@{}l|c|c|c@{}}
      \textbf{CIFAR-10} &
         \begin{tabular}[x]{@{}c@{}}Params\\(M)\end{tabular} &
         \begin{tabular}[x]{@{}c@{}}Training Time\\(GPU days)\end{tabular} &
         \begin{tabular}[x]{@{}c@{}}Test Error\\(\%)\end{tabular}
         \\ \hline
      NASNet-A     &      3.3    &  1800        &     2.65 \\
      NASNet-A     &     27.6    &  1800        &     2.4  \\
      AmoebaNet-B  &      2.8    &  3150        &     2.55 \\
      AmoebaNet-B  &     13.7    &  3150        &     2.31 \\
      AmoebaNet-B  &     26.7    &  3150        &     2.21 \\
      AmoebaNet-B  &     34.9    &  3150        &     2.13 \\
      DARTS        &      3.4    &     4        &     2.83 \\
      SNAS         &      2.8    &     1.5      &     2.85 \\
      ENAS         &      4.6    &     0.45      &     2.89 \\
      \hline

      \vspace{-4pt}WRN 28-10              & \multirow{2}{*}{36.4} & \multirow{2}{*}{0.4} & \multirow{2}{*}{3.08} \\
      \scriptsize{(baseline with cutout)} &                       &                      &                       \\ \hline
      SWRN 28-4-2   &    2.7    &      0.12      &     3.45  \\
      SWRN 28-6-2   &    6.1    &      0.25      &     3.0  \\
      \hdashline
      SWRN 28-10   &     36.4    &     0.4      &     2.7  \\
      SWRN 28-10-2 &     17.1    &     0.4      &     2.69 \\
      \hdashline
      SWRN 28-14   &     71.4    &     0.7      &     2.55 \\
      SWRN 28-14-2 &     33.5    &     0.7      &     2.53 \\
   \end{tabular}
}
\end{table}

%% file: include/hcnn/fig_swrn_folding.tex
\begin{figure}
    \centering
    \includegraphics[trim={0 0.3cm 0cm 0cm},clip,width=0.85\textwidth]{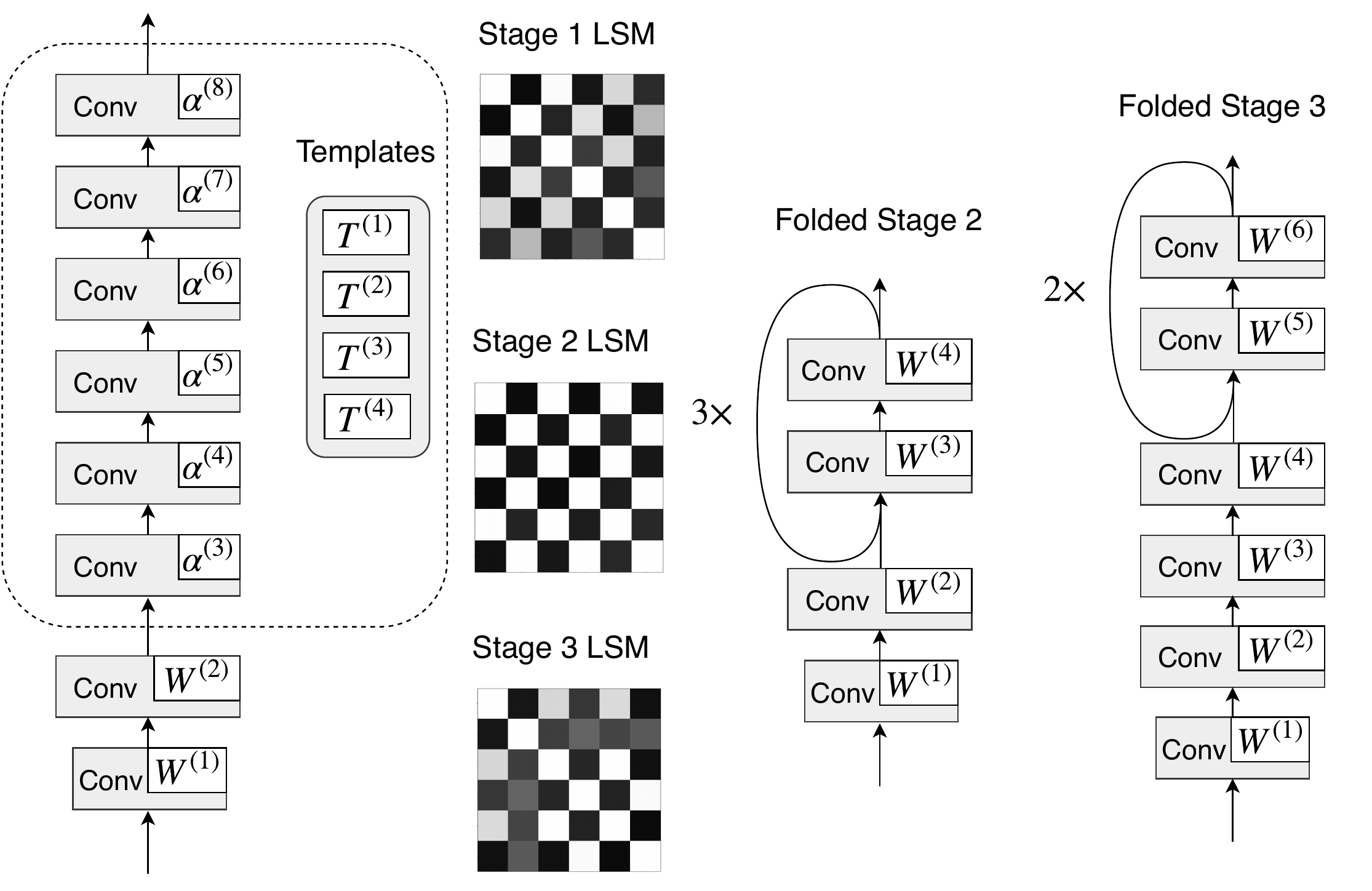}
    \caption{
      Extracting implicit recurrences from a SWRN 28-10-4.
      \emph{\textbf{Left:}}
         Illustration of the stages of a SWRN-28-10-4 (residual connections
         omitted for clarity).  The first two layers contain individual
         parameter sets, while the other six share four templates. All 3 stages
         of the network follow this structure.
      \emph{\textbf{Middle:}}
         LSM for each stage after training on CIFAR-10, with many elements
         close to $1$.  Hard sharing schemes can be created for pairs with
         large similarity by tying their coefficients (or, equivalently, their
         effective weights).
      \emph{\textbf{Right:}}
         Folding stages 2 and 3 leads to self-loops and a CNN with recurrent
         connections -- LSM for stage 2 is a repetition of 2 rows/columns, and
         folding decreases the number of parameters.
      }
    \label{fig:swrn_folding}
\end{figure}

%% file: include/hcnn/fig_shortest.tex
\begin{figure}
\centering
\begin{subfigure}{.45\textwidth}
   \centering
   \includegraphics[width=0.65\linewidth]{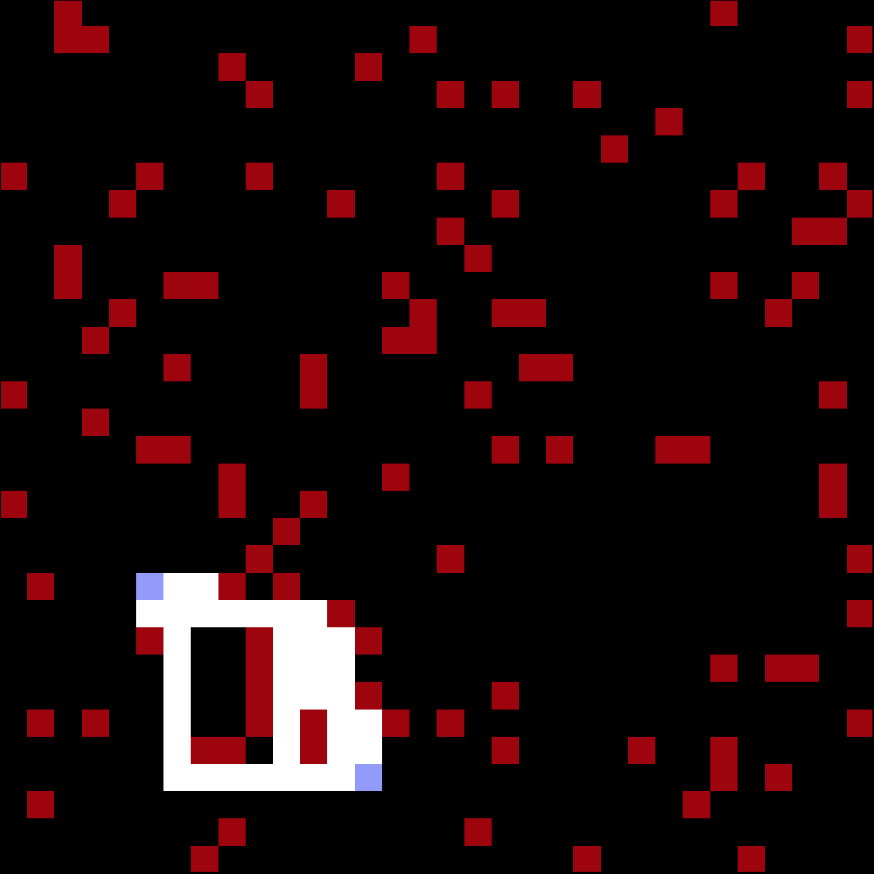}
   \caption{
      Generated example for the synthetic shortest paths task.  Blue pixels
      indicate the query points; red pixels represent obstacles, and white
      pixels are points in a shortest path (in terms of Manhattan distance)
      between query pixels. The task consists of predicting the white pixels
      (shortest paths) from the blue and red ones (queries and obstacles).
   }
  \label{fig:sub1}
\end{subfigure}%
\qquad
\begin{subfigure}{.45\textwidth}
   \centering
   \includegraphics[width=0.95\linewidth]{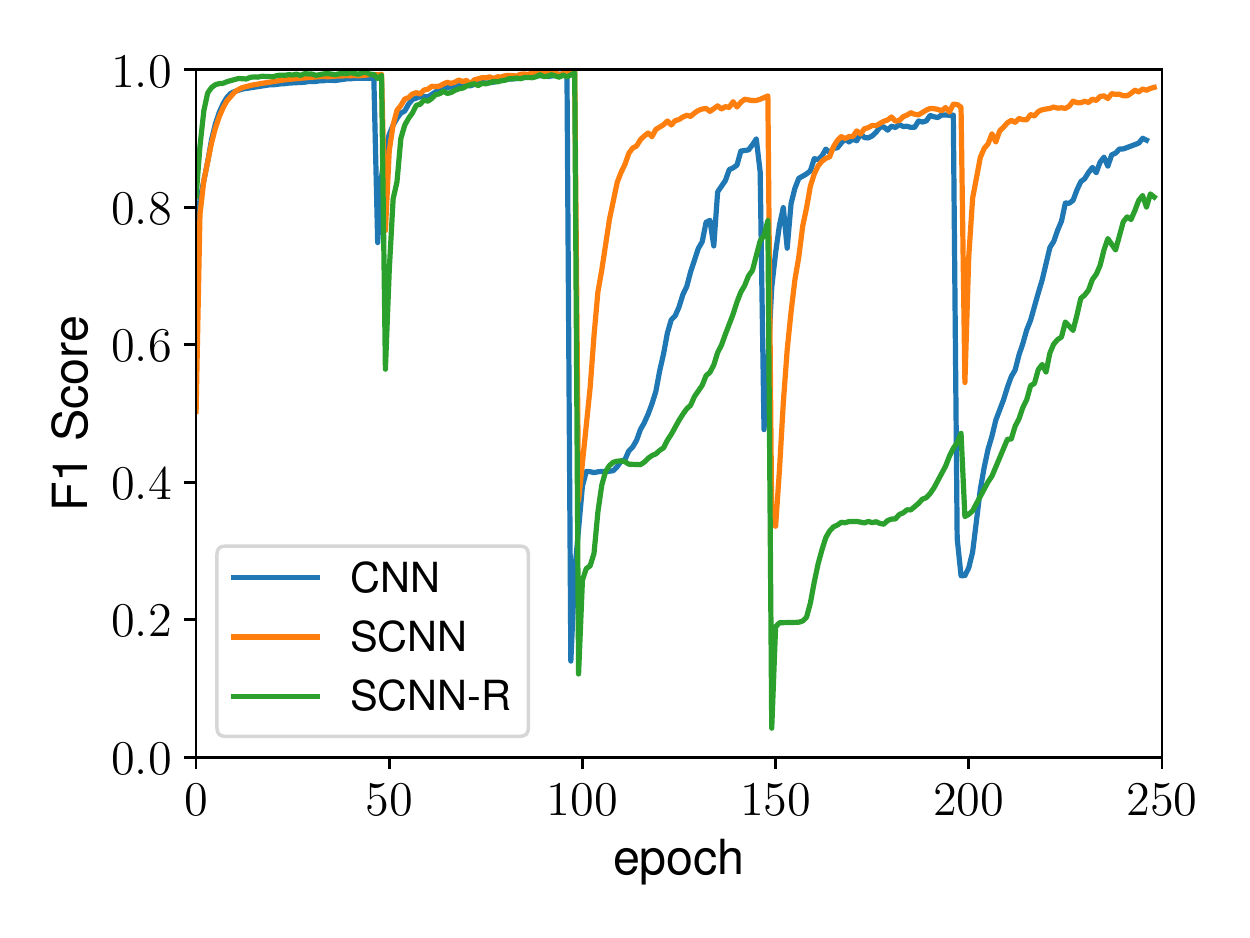}
   \vspace{-5pt}
   \caption{
      Training curves for the shortest paths task, where difficulty of
      examples increases every 50 epochs.  A SCNN adapts faster than a CNN to
      new phases and performs better, suggesting better extrapolation
      capacity.  With a recurrence regularizer $\lambda_R = 0.01$ (SCNN-R),
      the model makes faster progress on the first phase, but fails to adapt
      to harder examples.
   }
   \label{fig:sub2}
\end{subfigure}
\caption{Shortest paths task. Best viewed in color.}
\label{fig:shortest}
\end{figure}

%% file: include/hcnn/sec5_conclusion.tex
\section{Conclusion}
\label{sec:conclusion}

In this work, we take a step toward more modular and compact CNNs by extracting
recurrences from feed-forward models where parameters are shared among layers.
Experimentally, parameter sharing yields models with lower error on CIFAR and
ImageNet, and can be used for parameter reduction by training in a regime with
fewer parameter templates than layers.  Moreover, we observe that parameter
sharing often leads to different layers being functionally equivalent after
training, enabling us to collapse them into recurrent blocks.  Results on an
algorithmic task suggest that our shared parameter structure beneficially
biases extrapolation.  We gain a more flexible form of behavior typically
attributed to RNNs, as our networks adapt better to out-of-domain examples.
Our form of architecture discovery is also competitive with neural
architecture search (NAS) algorithms, while having a smaller training cost
than state-of-the-art gradient-based NAS.

As the only requirement for our method is for a network to have groups of
layers with matching parameter sizes, it can be applied to a plethora of CNN
model families, making it a general technique with negligible computational
cost.  We hope to raise questions regarding the rigid definitions of CNNs and
RNNs, and increase interest in models that fall between these definitions.
Adapting our method for models with non-uniform layer parameter
sizes~\citep{densenet,sparsenet} might be of particular future interest.

%% file: include/hcnn/appendix.tex
\newpage
\appendix

\part*{Appendix}

\section{Additional Results for Implicit Recurrences} \label{appendix}

Section~\ref{sec-imp_rec} presents an example of implicit recurrences and
folding of a SWRN 28-10-4 trained on CIFAR-10, where, for example, the last 6
layers in the second stage of the network fold into 2 layers with a self-loop.

Figure~\ref{fig:appendix_ex1} presents an additional example, where non-trivial
recurrences (unlike the one in Figure~\ref{fig:swrn_folding}) emerge naturally,
resulting in a model that is rich in structure.

\input{include/hcnn/fig_appendix_ex1}

\input{include/hcnn/fig_appendix_several}

\input{include/hcnn/fig_appendix_evolution}

\section{Initialization of Coefficients} \label{appendix2}

During our initial experiments, we explored different initializations for the
coefficients $\boldsymbol{\alpha}$ of each layer, and observed that using an
orthogonal initialization \citep{ortho} resulted in superior performance
compared to uniform or normal initialization schemes.

Denote $\mA$ as the $L \times k$ matrix ($L$ is the number of layers sharing
parameters and $k$ the number of templates) with each $i$'th row containing
the coefficient of the $i$'th layer $\boldsymbol{\alpha}^{(i)}$.  We initialize
it such that $\mA^T \mA = \mI$, leading to
$\forall_{i},
   \langle \boldsymbol{\alpha}^{(i)}, \boldsymbol{\alpha}^{(i)} \rangle = 1$
and
$\forall_{i \neq j},
   \langle \boldsymbol{\alpha}^{(i)}, \boldsymbol{\alpha}^{(j)} \rangle = 0$.
While our choice for this is mostly empirical, we believe that there is likely
a connection with the motivation for using orthogonal initialization for RNNs.

Moreover, we discovered that other initialization options for $\mA$ work
similarly to the orthogonal one.  More specifically, either initializing
$\mA$ with the identity matrix when $L=k$ (which naturally leads to
$\mA^T \mA = \mI$) or enforcing some sparsity (initialize $\mA$ with a uniform
or normal distribution and randomly setting half of its entries to zero)
performs similarly to the orthogonal initialization in a consistent manner.
We believe the sparse initialization to be the simplest one, as each
coefficient $\boldsymbol{\alpha}$ can be initialized independently.

Finally, note that having $\mA^T \mA = \mI$ results in the Layer Similarity
Matrix also being the identity at initialization (check that
$S_{i,j} =
   \frac{|\langle \boldsymbol{\alpha}^{(i)}, \boldsymbol{\alpha}^{(j)} \rangle|}
        {\lVert \boldsymbol{\alpha}^{(i)} \rVert \lVert \boldsymbol{\alpha}^{(j)} \rVert}
=  \frac{|(\mA^T \mA)_{i,j}|}
        {\lVert \boldsymbol{\alpha}^{(i)} \rVert \lVert \boldsymbol{\alpha}^{(j)} \rVert}$,
so if $(\mA^T \mA)_{i,j} = 1$, then $S_{i,j} = 1$, and the same holds for $0$.
Surprisingly, even though the orthogonal initialization leads to a LSM that has
no structure in the beginning of training, the rich patterns that we observe
still emerge naturally after optimization.

%% file: include/hcnn/fig_appendix_ex1.tex
\begin{figure}[h]
   \centering
   \includegraphics[width=0.8\textwidth]{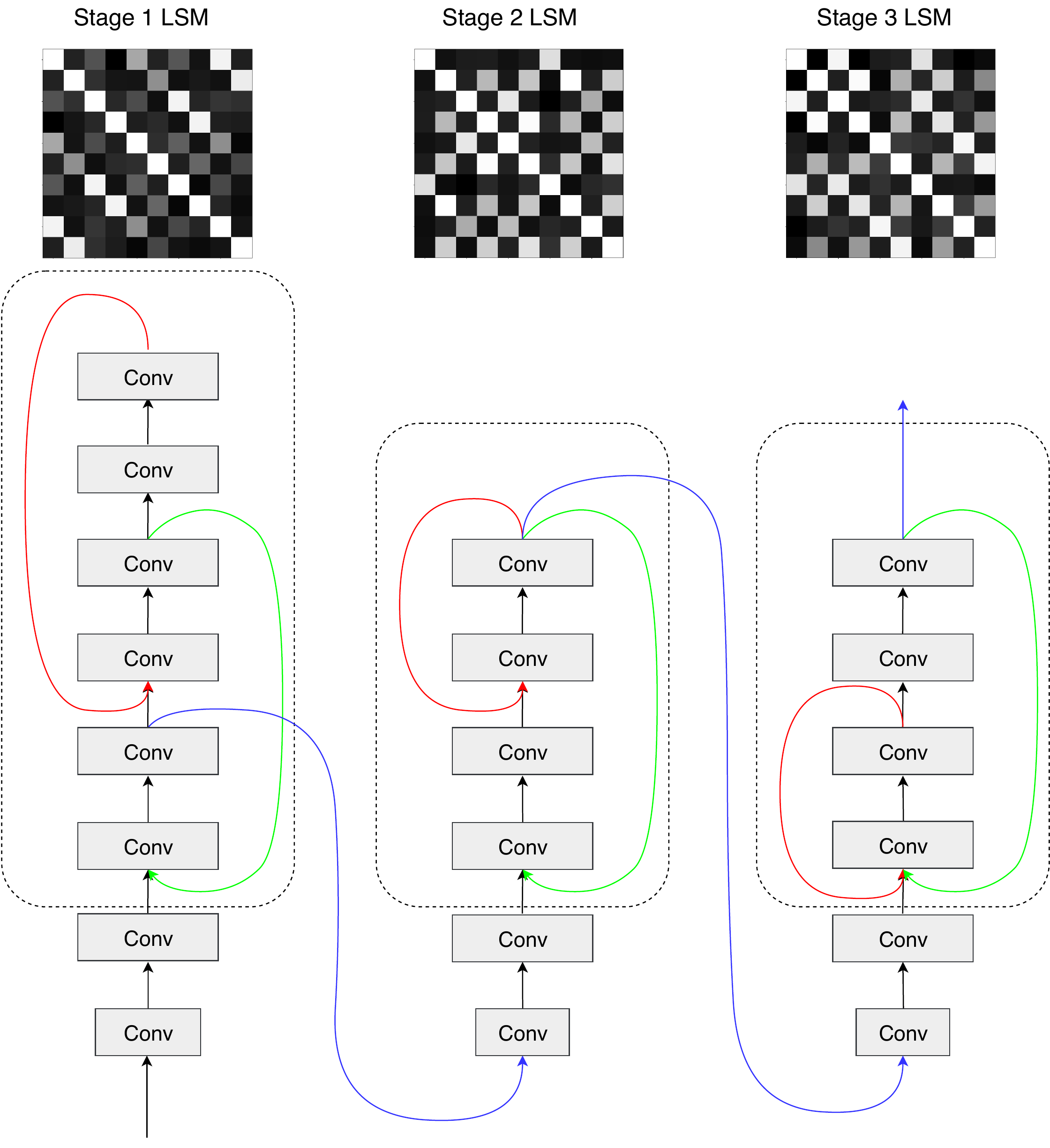}
   \caption{
      SWRN 40-8-8 (8 parameter templates shared among groups of
      $\frac{40-4}{3} -2 = 10$ layers) trained with soft parameter sharing on
      CIFAR-10.  Each stage (originally with 12 layers -- the first two do not
      participate in parameter sharing) can be folded to yield blocks with
      complex recurrences.  For clarity, we use colors to indicate the
      computational flow: red takes precedence over green, which in turn has
      precedence over blue.  Colored paths are only taken once per stage.
      Although not trivial to see, recurrences in each stage's folded form are
      determined by row/column repetitions in the respective Layer Similarity
      Matrix.  For example, for stage 2 we have
         $S_{5,3} \approx S_{6,4} \approx 1$,
      meaning that layers 3, 4, 5 and 6 can be folded into layers 3 and 4 with
      a loop (captured by the red edge).  The same holds for
      $S_{7,1}$, $S_{8,2}$, $S_{9,3}$ and $S_{10,4}$, hence after the loop
      with layers 3 and 4, the flow returns to layer 1 and goes all the way to
      layer 4, which generates the stage's output.  Even though there is an
      approximation when folding the network (in this example, we are tying
      layers with similarity close to $0.8$), the impact on the test error is
      less than $0.3\%$.  Also note that the folded model has a total of 24
      layers (20 in the stage diagrams, plus 4 which are not shown,
      corresponding to the first layer of the network and three $1 \times 1$
      convolutions in skip-connections), instead of the original 40.
   }
   \label{fig:appendix_ex1}
\end{figure}

%% file: include/hcnn/fig_appendix_several.tex
\begin{figure}[h]
   \centering
   \includegraphics[width=\textwidth]{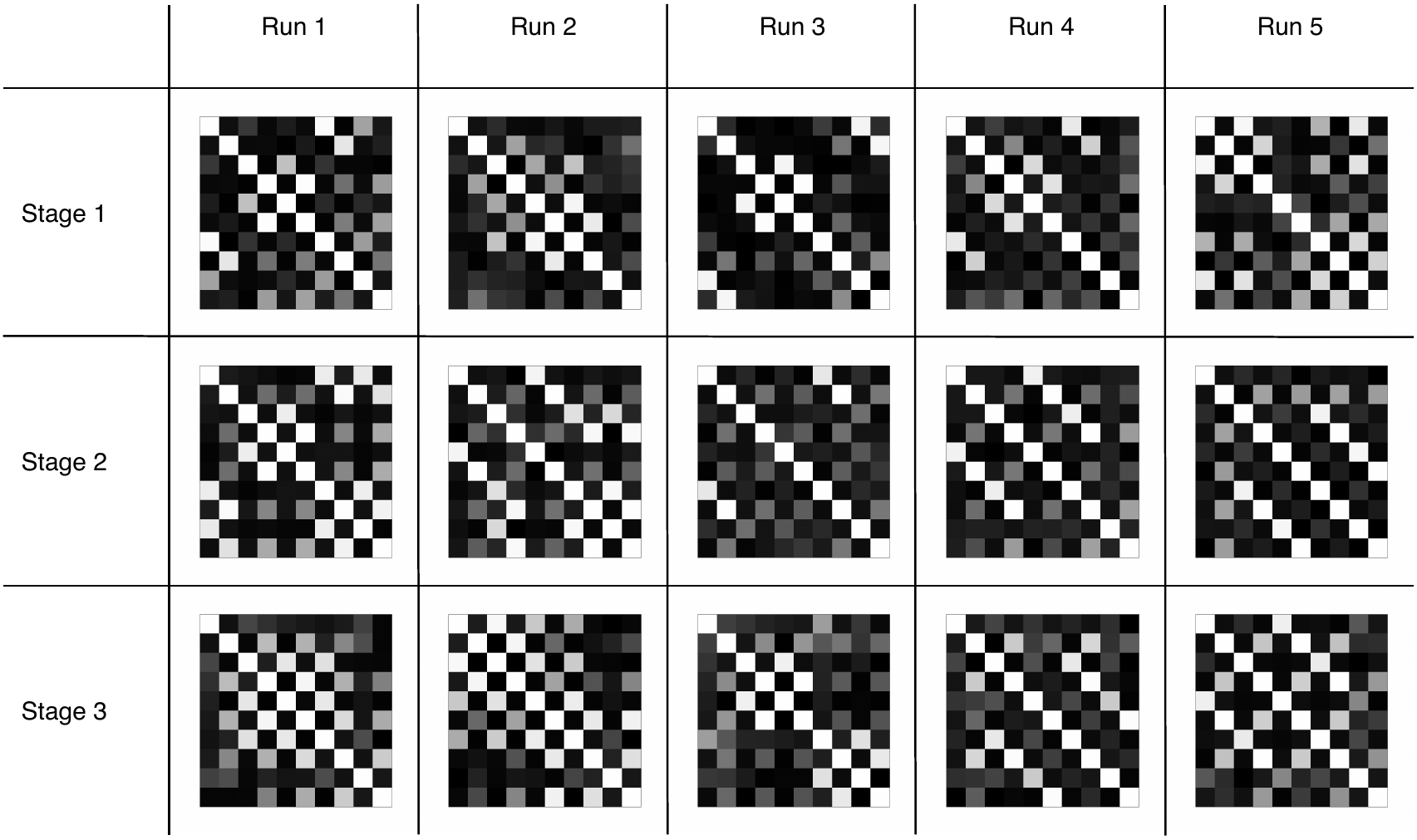}
   \caption{
      LSMs of a SWRN 40-8-8 (composed of 3 stages, each with 10 layers sharing
      8 templates) trained on CIFAR-10 for 5 runs with different random seeds.
      Although the LSMs differ across different runs, hard parameter sharing
      can be observed in all cases (off-diagonal elements close to 1, depicted
      by white), characterizing implicit recurrences which would enable
      network folding.  Moreover, the underlying structure is similar across
      runs, with hard sharing typically happening among layers $i$ and $i+2$,
      leading to a ``chessboard'' pattern.
   }
   \label{fig:appendix_several}
\end{figure}

%% file: include/hcnn/fig_appendix_evolution.tex
\begin{figure}[h]
   \centering
   \includegraphics[width=\textwidth]{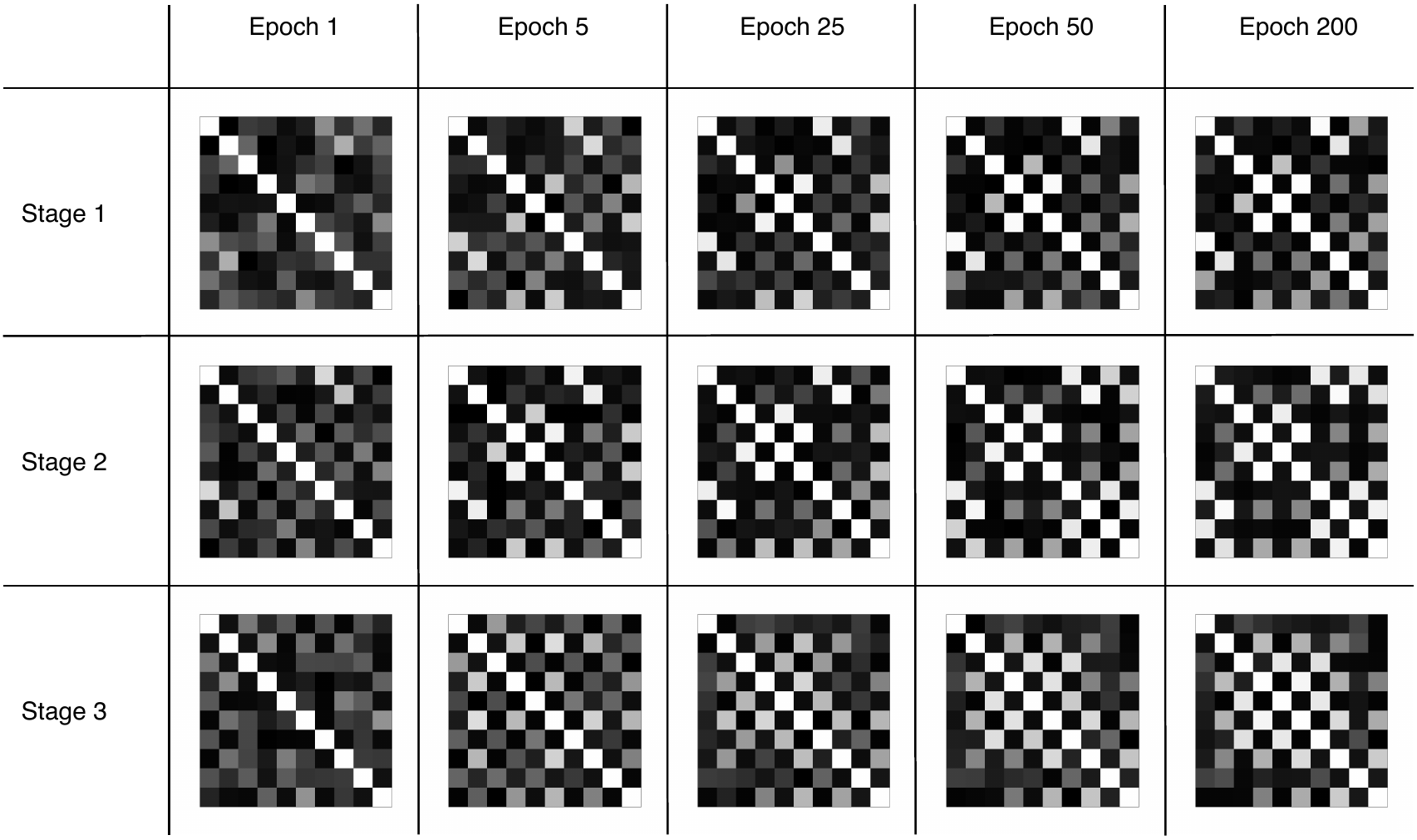}
   \caption{
      LSMs of a SWRN 40-8-8 (composed of 3 stages, each with 10 layers sharing
      8 templates) at different epochs during training on CIFAR-10.  The
      transition from an identity matrix to the final LSM happens mostly in
      the beginning of training: at epoch 50, the LSM is almost
      indistinguishable from the final LSM at epoch 200, and most of the final
      patterns are observable already at epoch 25.
   }
   \label{fig:appendix_evolution}
\end{figure}